\documentclass[letterpaper, 10 pt, conference]{ieeeconf}  

\IEEEoverridecommandlockouts                              

\usepackage{cite}
\usepackage{amsmath,amssymb,amsfonts}
\usepackage{graphicx}
\usepackage{svg}
\usepackage{textcomp}
\usepackage{xcolor}
\usepackage{algorithm}
\usepackage{algpseudocode}
\usepackage{booktabs}
\usepackage{glossaries}
\usepackage{subcaption}
\usepackage{float}

\usepackage{amsthm}
\theoremstyle{definition}
\newtheorem{defn}{Definition}

\usepackage{hyperref}

\title{\LARGE \bf
``Why This Avoidance Maneuver?'' Contrastive Explanations \\ in Human-Supervised Maritime Autonomous Navigation
}

\iftrue
\author{Joel Jose$^{1}$, Andreas Madsen$^{2}$, Andreas Brandsæter$^{3}$, Tor A. Johansen$^{4}$, and Erlend M. Coates$^{1}$%
\thanks{$^{1}$Department of ICT and Natural Sciences, 
Norwegian University of Science and Technology (NTNU), Ålesund, Norway. %
{\tt\small \{joel.j.chirapanath, erlend.coates\}@ntnu.no}%
}%
\thanks{$^{2}$Department of Ocean Operations and Civil Engineering, 
Norwegian University of Science and Technology (NTNU), Ålesund, Norway. %
{\tt\small andreas.madsen@ntnu.no}}%
\thanks{$^{3}$Department of Science and Mathematics, 
Volda University College (HVO), Volda, Norway. %
{\tt\small andreas.brandsaeter@hivolda.no}}%
\thanks{$^{4}$Department of Engineering Cybernetics, 
Norwegian University of Science and Technology (NTNU), Trondheim, Norway. %
{\tt\small tor.arne.johansen@ntnu.no}}%
}
\fi

\begin{document}

\maketitle
\thispagestyle{empty}
\pagestyle{empty}


\begin{abstract}
Automated maritime collision avoidance will rely on human supervision for the foreseeable future. This necessitates transparency into how the system perceives a scenario and plans a maneuver. However, the causal logic behind avoidance maneuvers is often complex and difficult to convey to a navigator. This paper explores how to explain these factors in a selective, understandable manner for supervisors with a nautical background. We propose a method for generating contrastive explanations, which provide human-centric insights by comparing a system's proposed solution against relevant alternatives. To evaluate this, we developed a framework that uses visual and textual cues to highlight key objectives from a state-of-the-art collision avoidance system. An exploratory user study with four experienced marine officers suggests that contrastive explanations support the understanding of the system's objectives. However, our findings also reveal that while these explanations are highly valuable in complex multi-vessel encounters, they can increase cognitive workload, suggesting that future maritime interfaces may benefit most from demand-driven or scenario-specific explanation strategies.




\end{abstract}


\section{Introduction}

Employing maritime autonomous surface ships (MASS) in the ocean space can potentially reduce operating costs, increase efficiency, and lower casualties arising from human errors, while also compensating for increasing global demand for certified officers \cite{kurt_asvimpacts_2022}. However, the complexity in perception and dynamics of the maritime environment, coupled with maritime rules and regulations such as the Convention on the International Regulations for Preventing Collision at Sea (COLREGs), makes automated collision avoidance at sea a challenging planning and decision-making problem \cite{vagale_path_2021}. These challenges prevent MASS from operating fully unsupervised in the near future, especially when at risk of collision or grounding, and therefore, warrant supervision through trained personnel who can ensure that maneuvers from a collision avoidance system (CAS) follow safety and regulatory requirements \cite{rodseth_towards_2022}.


A human operator's primary role when supervising a CAS is to monitor performance and intervene if the system creates high-risk conditions\cite{madsen_improving_2025}. To do this effectively, the supervisor requires high situational awareness (SA) of the scenario and the system itself. Endsley \cite{endsley_supporting_2023} highlights two key methods for achieving this: \textbf{transparency} and \textbf{explainability}. Transparency involves sharing information about the system's perception and current actions, while explainability clarifies the underlying objectives and decision-making involved behind those actions. While recent maritime research has shown promise in utilizing transparency to develop human-machine-interfaces \cite{madsen_improving_2025, van_de_merwe_influence_2024} based on established SA models (\cite{endsley_toward_1995, psw_model}), we aim to extend this work by incorporating explainability to support the supervision of a CAS. Specifically, we seek to explain the behaviour and objectives behind automated maneuvers in a meaningful, focused, and understandable manner for supervisors with a nautical background.


\begin{figure}[t]
    \centering
    \includegraphics[width=1\linewidth]{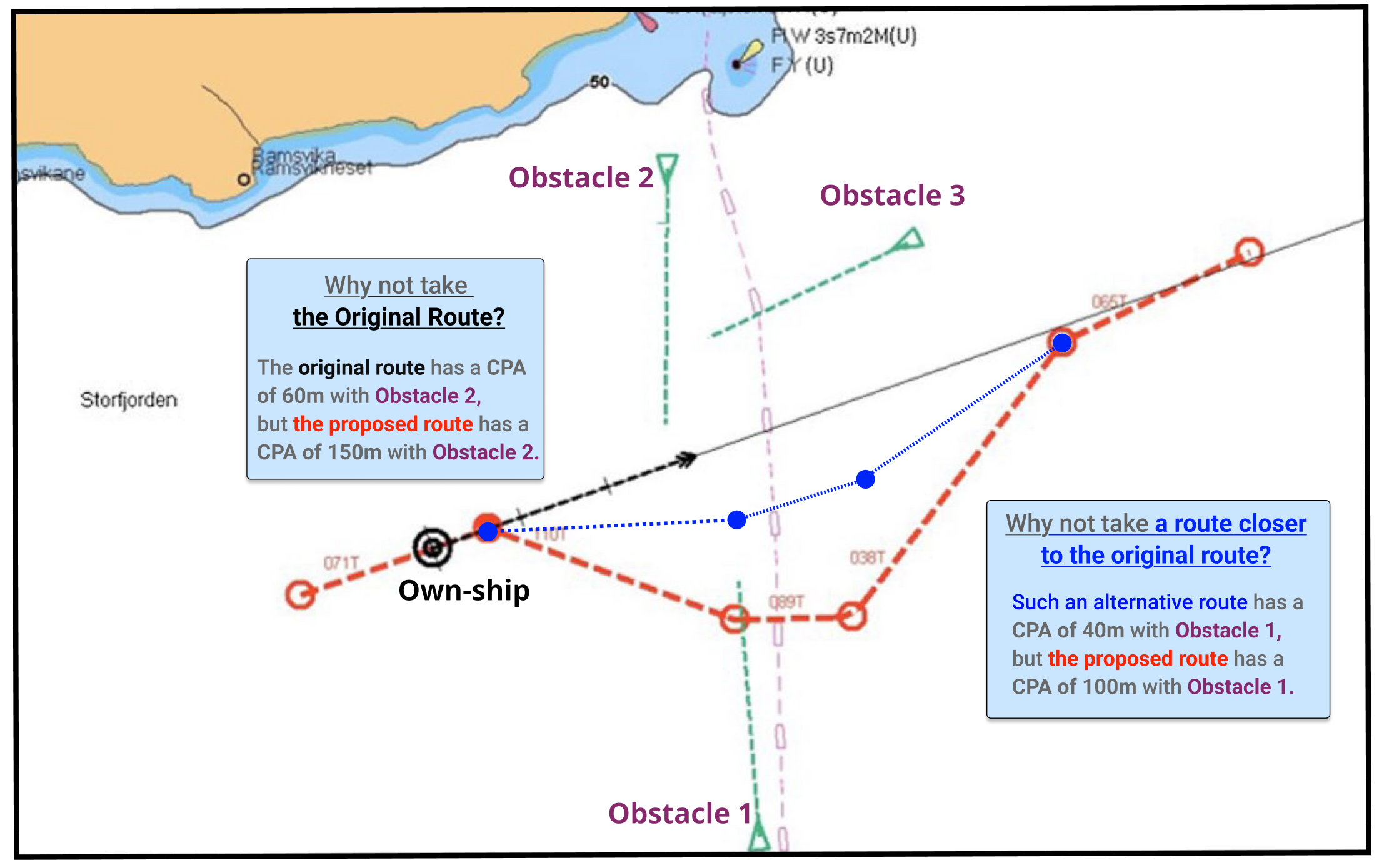}
    \caption{Concept illustration. Contrastive explanations for a collision avoidance maneuver visualized in an Electronic Chart Display and Information System (ECDIS) (Explanations visualized on top of a figure from \cite{brandsaeter_simulator-based_2024}). Here, a proposed maneuver (red) is compared with the original route (black), and an alternative route (blue) based on an objective of keeping a safe closest point of approach (CPA).
    }
    \label{fig:ecdis_illustration}
\end{figure}

Explainability is widely studied in various robotic applications such as autonomous driving\cite{ Omeiza2022ExplanationsSurvey}, robot manipulators \cite{gjaerum_real-time_2023}, and the maritime domain\cite{gjaerum_explaining_2021, singh_fuzzyxai2023, veitch_human-centered_2021} as a means of revealing the complex decision-making of intelligent systems in a manner understandable to humans.  Research in the social sciences indicates that humans typically ask questions of a contrastive nature when seeking an explanation for a complex event \cite{Lipton1990, miller_explanation_2019}. By answering a query such as ``\textit{Why does the CAS perform maneuver A instead of a maneuver B}'', we focus on a potential key factor behind a solution, instead of identifying all causal reasons that led to it. This can be achieved by highlighting how maneuver A better satisfies one or more objectives compared to maneuver B, as in a CAS that solves a multi-objective optimization problem \cite{vagale_path_2021}. Such explanations follow an option-centric rationale \cite{option_centric_Luo2019, gros_scaffolding_2023}, providing a means of interacting with the system and promoting trust and acceptance of the system's actions. Contrastive explanations, therefore, show promise as a way of generating human-centric insights for autonomous driving, scheduling, and motion planning for mobile robots\cite{krarup_contrastive_2021, omeiza_towards_2021, brandao_towards_2021, schneider_ce-mrs_2024}.


In this study, we examine the usefulness of contrastive explanations for understanding the underlying objectives of a maritime Collision Avoidance System (CAS) and, in turn, their impact on enabling effective and satisfactory human supervision of such a system. Specifically, this paper makes the following contributions:

\begin{itemize}
    \item \textbf{Explanation Generation:}  
    We propose a method for producing contrastive explanations that highlight causal features in the CAS's cost function and allow supervisors to compare alternative trajectories.

    \item \textbf{Visualization Interface:}  
    We develop a framework for generating contrastive explanations from a widely used collision avoidance planner and design a visualization interface that presents transparency information and contrastive explanations for possible maneuvers.

    \item \textbf{Exploratory User Study:}  
    We conduct an exploratory study investigating how experienced seafarers perceive the utility and satisfaction of receiving contrastive explanations when supervising a CAS. Through this, we investigate the following research questions:

\begin{enumerate}
    \item Do experienced seafarers supervising a CAS find contrastive explanations helpful for understanding the system's underlying objectives?
    \item Do they find the availability of alternative trajectories—paired with contrastive explanations—useful?
    \item Are they better able to determine whether to accept or reject a proposed maneuver when such contrastive explanations are provided?
\end{enumerate}
\end{itemize}

The remainder of this paper is structured as follows: We formalize the problem in \autoref{section:preliminaries}, along with the specifications of the CAS we have used. We then introduce a framework for generating contrastive explanations for maritime collision avoidance in \autoref{section:ce_framework}. Subsequently, \autoref{section:user_study} outlines the exploratory user study conducted to obtain qualitative insights and feedback, after which \autoref{section:results} presents the findings from the user study, and \autoref{section:conclusions} concludes with directions for future work.
\section{Problem Definition}\label{section:preliminaries}

In this section, we formalize the problem of collision avoidance and the supervision of a CAS, and conclude with specifications of the type of CAS used in this study.

\subsection{Collision Avoidance}

We consider collision avoidance as a multi-objective optimization problem \cite{vagale_path_2021}, where a CAS is employed on a ship (referred to as the ownship) for a situation $S$ involving the risk of collision or grounding. The CAS searches for collision-free trajectories that are predictable, compliant with COLREGs, and close to a ship's nominal trajectory (often retrieved from a global/mission planner). For each trajectory $\tau\in \mathcal{T}$, where $\mathcal{T}$ is the set of trajectories representing the search space of the CAS, objectives such as risk reduction, operational requirements, and efficiency are evaluated using a cost function $\phi(S,\tau)$. Each trajectory $\tau\in \mathcal{T}$ is generated by approximating the dynamics of the ownship. In addition, the motion of obstacles is predicted using heuristic, scenario-based, or collaborative methods \cite{AKDAG2022}. Thus, the role of the CAS is to identify a solution trajectory $\tau^*$ for the ownship as follows:
\begin{equation}
    \tau^* = \operatorname*{arg min}_{\tau \in \mathcal{T}} \phi(S,\tau)
\end{equation}

In this study, we consider the cost function $\phi(S,\tau)$ to be linearly additive, such that $\phi(S,\tau) = \sum_{f\in F}f(S,\tau)$. Here, each component $f \in F$ represents a distinct, human-interpretable objective (e.g., collision risk, efficiency). This decomposition allows us to identify causal factors—or ``features''—that drive the system's evaluation of trajectories.


\subsection{Supervision of a CAS}
We consider the scenario in which a human supervisor is alerted when the CAS detects that a situation $S$ has begun. The CAS proposes a solution trajectory $\tau^*$ to the operator for execution. The supervisor's role is to assess the information provided by the CAS, and either (1) approve the solution trajectory for execution, or (2) take over control of the vessel.

\begin{figure}[h]
    \centering
    \includegraphics[width=1\linewidth]{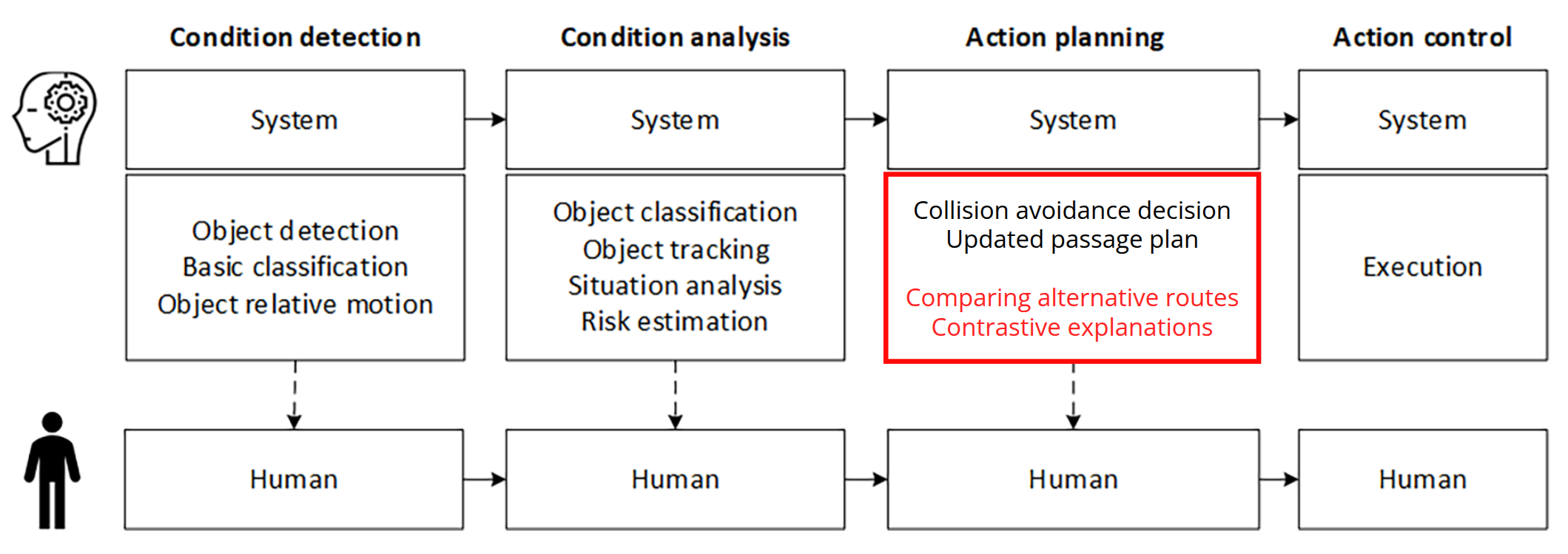}
    \caption{The adapted PSW model from \cite{van_de_merwe_towards2023}. We believe that comparing alternative routes with contrastive explanations fits best in supporting supervision of the action planning stage (highlighted in red).}
    \label{fig:psw_model}
\end{figure}

Van de Merwe et al. \cite{van_de_merwe_towards2023} proposed an adapted version of the Parasuraman, Sheridan, and Wickens (PSW) model \cite{psw_model} to categorize the task of collision avoidance of a CAS into unique information processing stages. Furthermore, the adapted model is utilized to propose transparency layers that can reveal insight into each of the stages \cite{van_de_merwe_influence_2024}, thereby supporting the supervisor's understanding of the CAS's functions. Following the adapted PSW model \cite{van_de_merwe_towards2023}, the CAS's action-planning stage involves selecting the solution trajectory $\tau^*$ from available alternatives. Therefore, we deduce that contrastive explanations—which clarify why $\tau^*$ was chosen over a specific alternative—directly enhance a supervisor's understanding of this critical stage (\autoref{fig:psw_model}).

\subsection{Contrastive Explanations for Collision Avoidance} \label{section:CE_def}
Following established terminology \cite{miller_explanation_2019, schneider_ce-mrs_2024}, generating a contrastive explanation for a situation $S$ requires comparing the system's proposed solution trajectory $\tau^*$ (the fact) against a supervisor-specified alternative $\tau'$ (the foil).
We assume that $\tau' \in \mathcal{T}$, which implies that $\phi(S,\tau^*)\leq\phi(S,\tau')$. We define a contrastive explanation as follows: 

\begin{defn}
Given a fact $\tau^*$, a foil $\tau' \neq\tau^*$, and a cost function $\phi(S,\tau)$ composed of a set of additive components $F$, a contrastive explanation $\mathcal{E_S}$ utilizes the subset of components $F_{exp} \in F$ in order to reason why $\tau^*$ is chosen over $\tau'$, where $F_{exp}$ is given by: 
\begin{equation}
F_{exp} = \{ f \in F \mid f(S,\tau^*) < f(S,\tau')\}    
\end{equation}
\end{defn}

\subsection{Collision Avoidance System}
For our study, we implement the framework for the Simulation-based MPC (SB-MPC) planner by Johansen et al. \cite{johansen_ship_2016}, chosen for its prominence in maritime applications, and clear attribution of individual costs in its cost function. The SB-MPC planner generates each candidate trajectory $\tau \in \mathcal{T}$ by applying a constant speed $(U^\tau)$ and course $(\chi^\tau)$ offset from the references of an autopilot. In a receding horizon fashion, the ownship tracks a 5-second segment of the solution trajectory $\tau^*$, before the algorithm reruns. To improve predictive realism, we include the return-to-path implementation from \cite{hagen_scenario-based_2022}. Although these trajectories represent immediate decision steps rather than complex long-term maneuvers, they offer the best transparency into the system's current causal logic.



The cost function $\phi$ for the SB-MPC is composed of seven distinct costs, summarized in \autoref{tab:cost_functions}.
\begin{table}[htbp]
\centering
\caption{SB-MPC Cost Function Components. }
\label{tab:cost_functions}
\begin{tabular}{ll}
\toprule
\textbf{Cost} & \textbf{Description} \\ \midrule
$f_{\mathrm{DO}}$ & Collision risk with dynamic obstacles \\
$f_{\mathrm{CLRG}}, f_\mathrm{{Trans}}$ & COLREG compliance and transition costs \\
$f_{U^\mathrm{ref}}, f_{\chi^\mathrm{ref}}$ & Deviation from autopilot speed/course references \\
$f_{\Delta U}, f_{\Delta \chi}$ & Variation from previous speed/course states \\ \bottomrule
\end{tabular}
\end{table}

\section{Proposed Framework}\label{section:ce_framework}
\begin{figure} [tbh]
    \centering
    \includegraphics[width=1\linewidth]{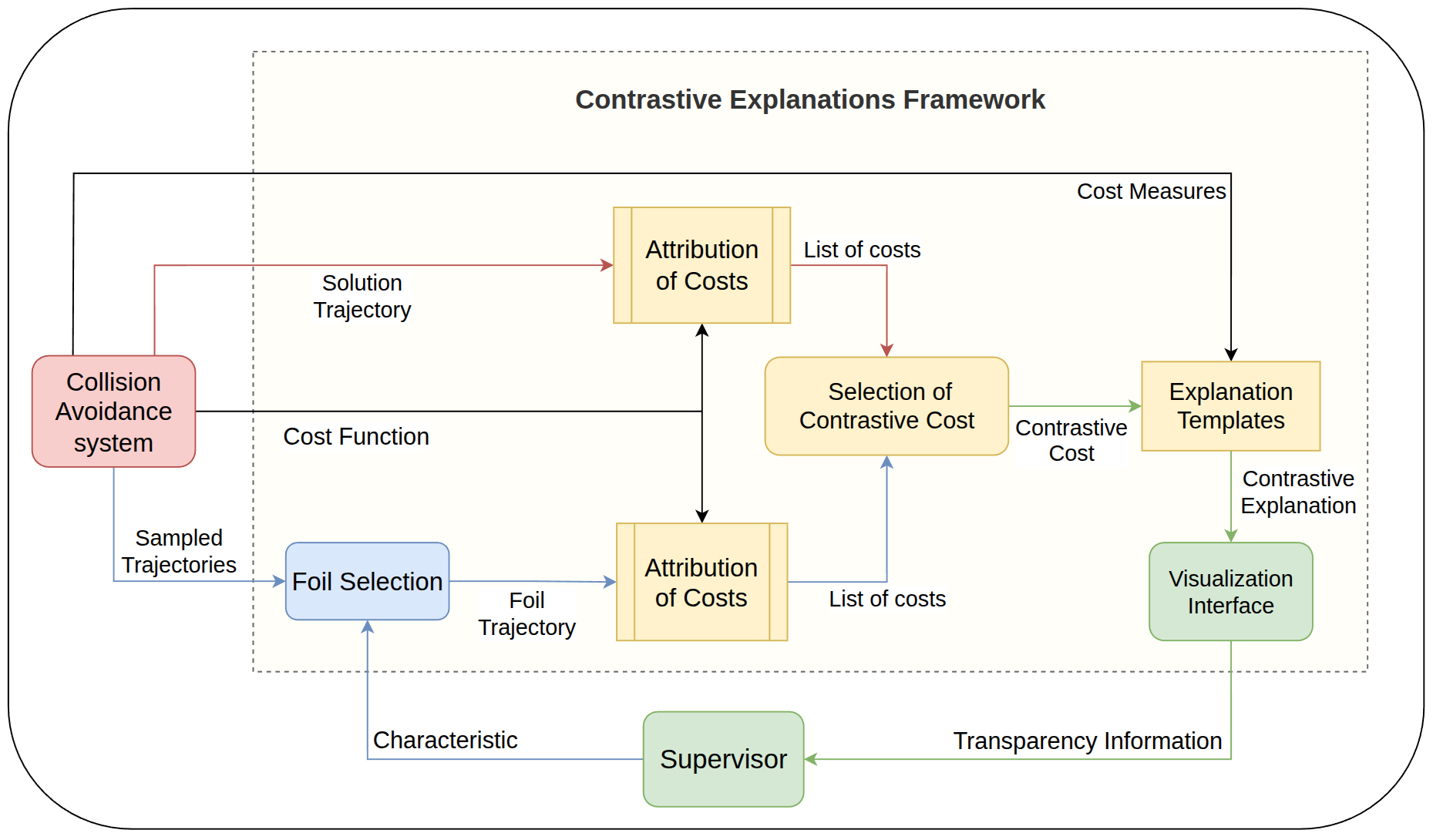}
    \caption{Contrastive Explanations Framework. The costs are extracted from the cost function for the solution and foil trajectories and compared to generate a contrastive explanation, which is visualized on an interface to the supervisor.} \label{fig:ce_framework}
\end{figure}

We introduce a general framework for contrastive explanations for CAS, as shown in \autoref{fig:ce_framework}. Our framework is a post-hoc tool \cite{heuillet_explainability_2021}, meaning that the explanations are generated after the collision avoidance planner has found a solution.

\subsection{Foil Selection}\label{section:traj_gen}

The utility of a contrastive explanation heavily depends on the choice of the foil, as the goal is to present an alternative that aligns with the operator's expectations. In this paper, we utilize the candidate trajectories generated by the SB-MPC as our source for maneuvers to be compared. 

We formulate explanations by comparing the fact $\tau^*$ with two foils: the \textbf{nominal} $\tau^{\mathrm{nominal}}$ (which follows the autopilot references), and the \textbf{alternative} $\tau^{\mathrm{alt}}$. We choose the alternative $\tau^{\mathrm{alt}}$ by filtering the search space $\mathcal{T}$ for a specific maneuver 'characteristic' (e.g., Port Turn, Reduced Speed) by imposing a corresponding constraint C defined in \autoref{table:characteristics}. This design choice aligns with how navigators communicate maneuvers \cite{hodne_convAI} and was informed by recurring discussions with a certified marine officer. From the filtered set of trajectories $\mathcal{T}^{alt}$, we choose the trajectory with the minimum total cost as the alternative $\tau^{\mathrm{alt}}$:
\begin{equation}
      \tau^{\mathrm{alt}}= \operatorname*{argmin}_{\tau \in \mathcal{T}^{\mathrm{alt}}} [\phi(S,\tau)]
\end{equation}

While selecting the characteristic is often left to the supervisor to allow for personalization~\cite{schneider_ce-mrs_2024, brandao_towards_2021, krarup_contrastive_2021}, for simplicity, we handpick characteristics that are intuitive and likely to be chosen by a seafarer for a specific scenario.

\begin{table}[h!]
\renewcommand{\arraystretch}{1.5} 
\centering
\begin{tabular}{c c}
\textbf{Characteristic} & \textbf{Constraint ($C$)} \\
\hline
Reduced speed & $ U^{\tau^{\mathrm{alt}}} < U^{\tau^*}$ \\
Port Turn & $ \chi^{\tau^{\mathrm{alt}}} < 0$ \\
Starboard Turn & $ \chi^{\tau^{\mathrm{alt}}} > 0$ \\
Closer to original route & $ |\chi^{\tau^{\mathrm{alt}}}| < |\chi^{\tau^*}|$ \\
Farther from original route & $ |\chi^{\tau^{\mathrm{alt}}}| > |\chi^{\tau^*}|$ \\
\end{tabular}
\caption{Choices of the characteristic and associated constraint for the alternative.}\label{table:characteristics}
\end{table}

\begin{figure*}
    \centering
    \includegraphics[width=0.8\linewidth]{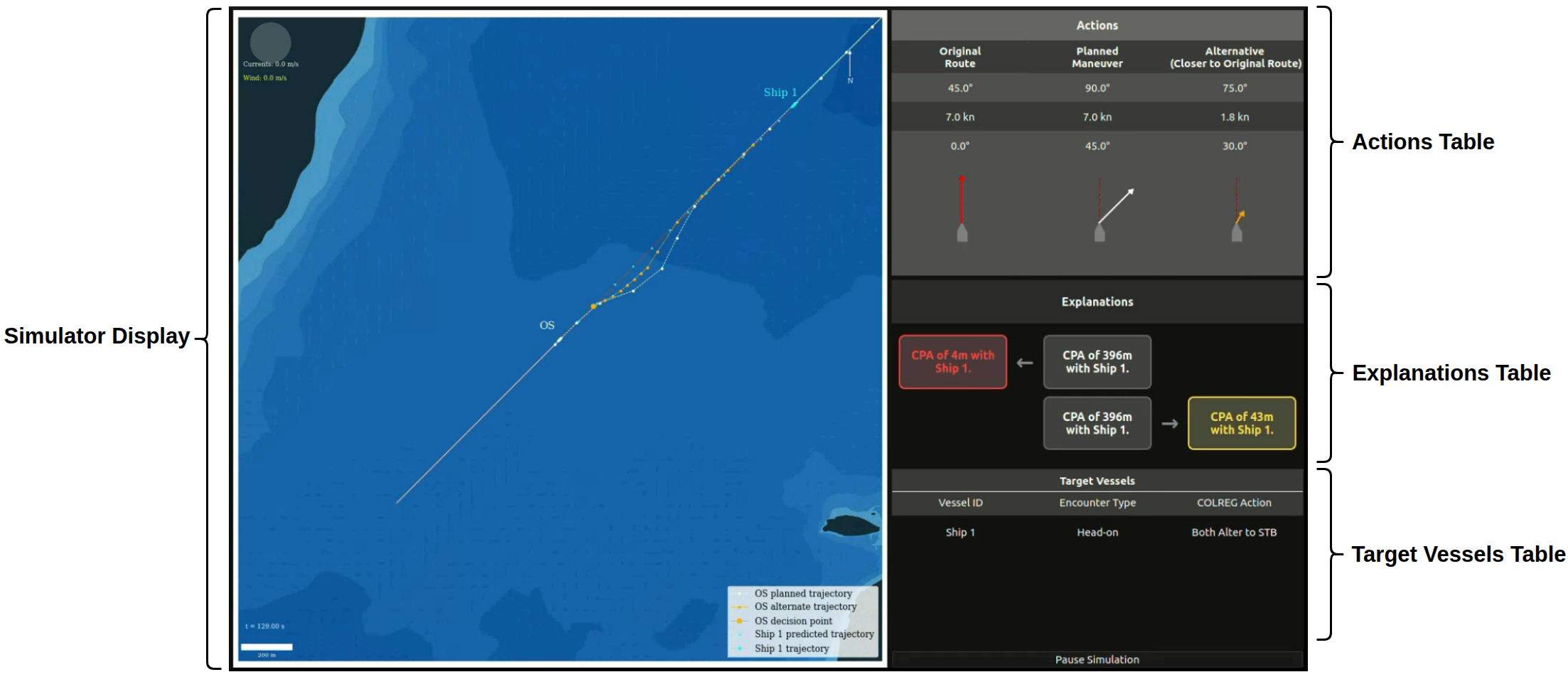}
    \caption{The Visualization Interface used in this study. The design draws inspiration from the transparency layers in \cite{van_de_merwe_influence_2024} and the explanations depicted in \autoref{fig:ecdis_illustration}. (Left) Map View illustrates the proposed fact (white), alternative (orange), with speed-proportional spacing of dots, and the nominal path (faint red). Obstacle predictions are shown as dots of the corresponding ship color. (Top right) Action table for all three trajectories. (Center right) Contrastive explanations comparing the fact against the foils. (Bottom right) Target vessel information}
    \label{fig:interface}
\end{figure*}

\subsection{Constructing the Explanation}\label{section:comparison}

Based on \autoref{section:CE_def}, we construct two pairs of contrastive explanations; one comparing the fact to the nominal, and another comparing the fact to the alternative corresponding to a characteristic. For each case, the explanations are constructed by following two objectives:

\subsubsection{Prioritized Information} 
To avoid overloading the user with information, we compare the fact with a foil using the objective for which the fact shows the greatest improvement over that foil. First, the subset of components $F_\mathrm{exp}$ is identified for a fact-foil pair according to \autoref{section:CE_def}. We then select the contrastive cost $f_{\mathrm{exp}} \in F_{\mathrm{exp}}$ for which the fact shows the greatest cost reduction relative to the foil:
\begin{equation}
    f_{\mathrm{exp}} = \operatorname*{argmin}_{f \in F_\mathrm{exp}}  [f(S,\tau^*) - f(S,\tau')],
\end{equation}

\noindent
where $\tau' \in \{\tau^{\mathrm{nominal}}, \tau^{\mathrm{alt}}\}$. 

\subsubsection{Semantic Cost Measures}
While our choice of the contrastive cost $f_{\mathrm{exp}}$ is based on its value attribution in $\phi$, the values themselves may not present sufficient meaning to a human supervisor monitoring the system \cite{veitch_human-centered_2021}. Therefore, we associate each cost with a corresponding 'cost measure' as shown in \autoref{tab:measure_mapping} that delegates meaning to the cost, while also being directly correlated to how the cost measure is calculated. We record these cost measures during the cost computation step of the SBMPC for each sampled trajectory.

\begin{table}[htbp]
\centering
\footnotesize 
\caption{Mapping of Costs to Cost Measures. (CPA refers to the closest point of approach between the ownship and the obstacle along their corresponding trajectories)}
\label{tab:measure_mapping}
\begin{tabular}{ll}
\toprule
\textbf{Cost} & \textbf{Associated Cost Measure} \\ \midrule
$f_{DO}$ & CPA distance to Dynamic Obstacle (DO) \\
$f_{CLRG}$ & COLREG rule number \\
$f_{Trans}$ & Transition behavior \\
$f_{U_{ref}}$ & Speed offset \\
$f_{\chi_{ref}}$ & Course offset \\ \bottomrule
\end{tabular}
\end{table}

Once the contrastive cost $f_{\mathrm{exp}}$ is identified, we map it to its corresponding measure in \autoref{tab:measure_mapping}. This measure is then used to populate a predefined text template, generating a concise contrastive explanation $\mathcal{E_S}$ that compares the fact and the foil as shown in \autoref{fig:interface}.


\subsection{Temporal Context and Triggers} \label{section:when_to_generate}
In this study, we evaluate the utility of contrastive explanations at some time before the decision point when a CAS initiates an avoidance maneuver, based on expert opinion and the simplicity of implementation. Since a supervisor needs time to process an explanation before responding to a situation, explanations are more effective when they target future events. To support this, we introduce two features in our explanation framework:

\subsubsection{Ahead-of-time Simulation}\label{section:aheadoftime}
We target explanations not towards actions that are currently executed, but future events that are predicted based on the ownship and obstacle vessel models. This is done by forward simulating the ownship (using the control behavior for $\tau^*$) and the obstacle states for 5 seconds, and rerunning the SB-MPC. This process is repeated until a time limit is reached, when the predicted states are less reliable or when an event triggers the generation of explanations, as described in the following section.

\subsubsection{Event Trigger}
 For each scenario, we generate an alternative trajectory and the contrastive explanations at the first instance where the SBMPC planner adds an offset to the autopilot references. Coupled with the ahead-of-time forward simulation described in \autoref{section:aheadoftime}, we provide explanations for a decision point occurring ahead of time.

\subsection{Visualization Interface}\label{section:interface}

We designed a visualization interface inspired by the transparency layers in \cite{van_de_merwe_influence_2024}, and integrated relevant information from the CAS onto the interface in real-time. As shown in Figure 3, we overlaid transparency information and explanations on the right side of the simulator's map view. The top-right table visualizes course and speed references for the fact, nominal, and alternative trajectories at a future decision point. Explanations are displayed in tinted boxes directly beneath their corresponding actions, allowing the supervisor to easily map the system's reasoning to the specific fact-foil pair. The target vessels table on the bottom right shows relevant information about what the ownship has perceived and understood about target vessels in the vicinity.



\begin{figure*}[ht] 
\centering
\begin{tabular}{cc} 
\includegraphics[width=0.45\textwidth]{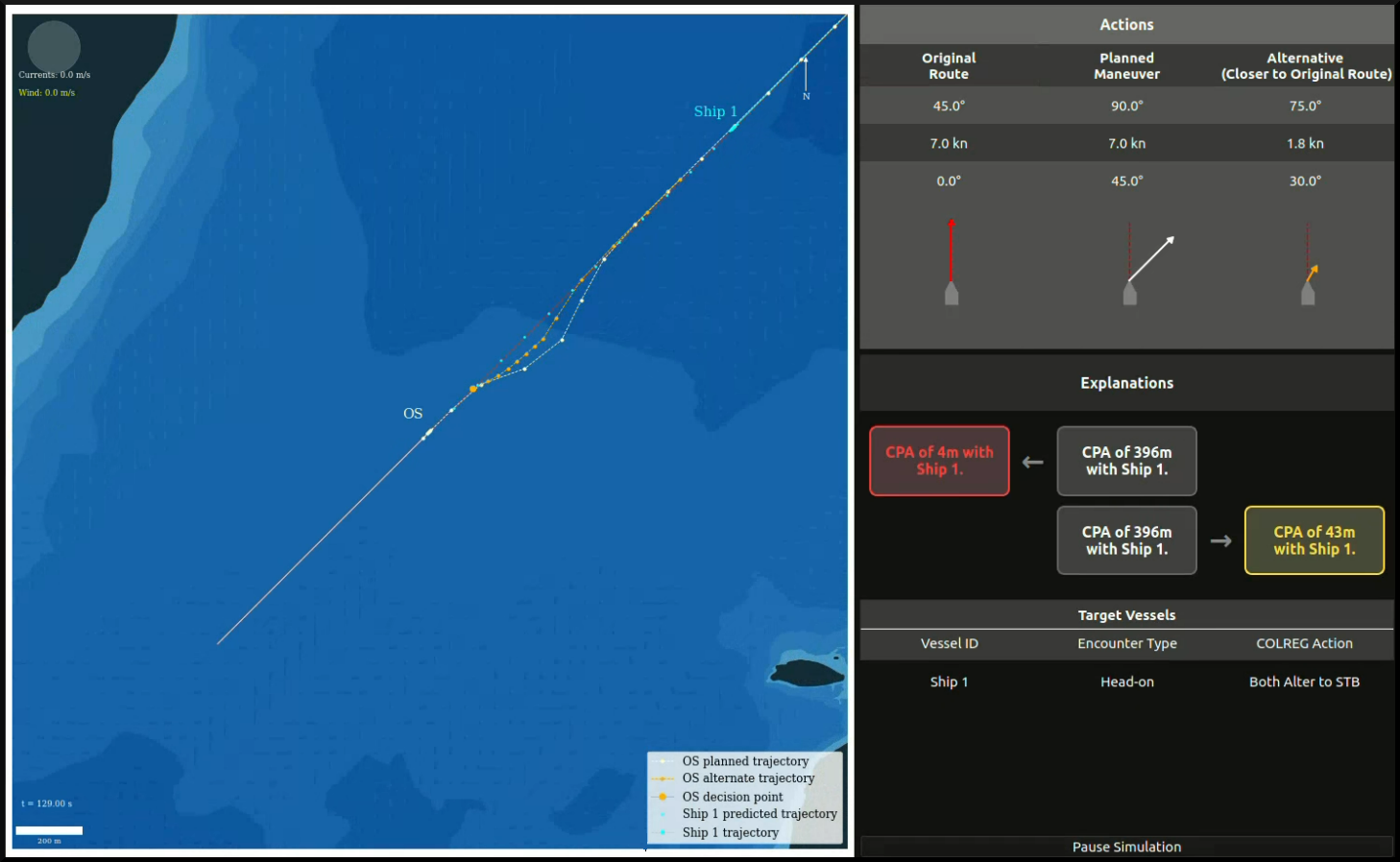} &
\includegraphics[width=0.45\textwidth]{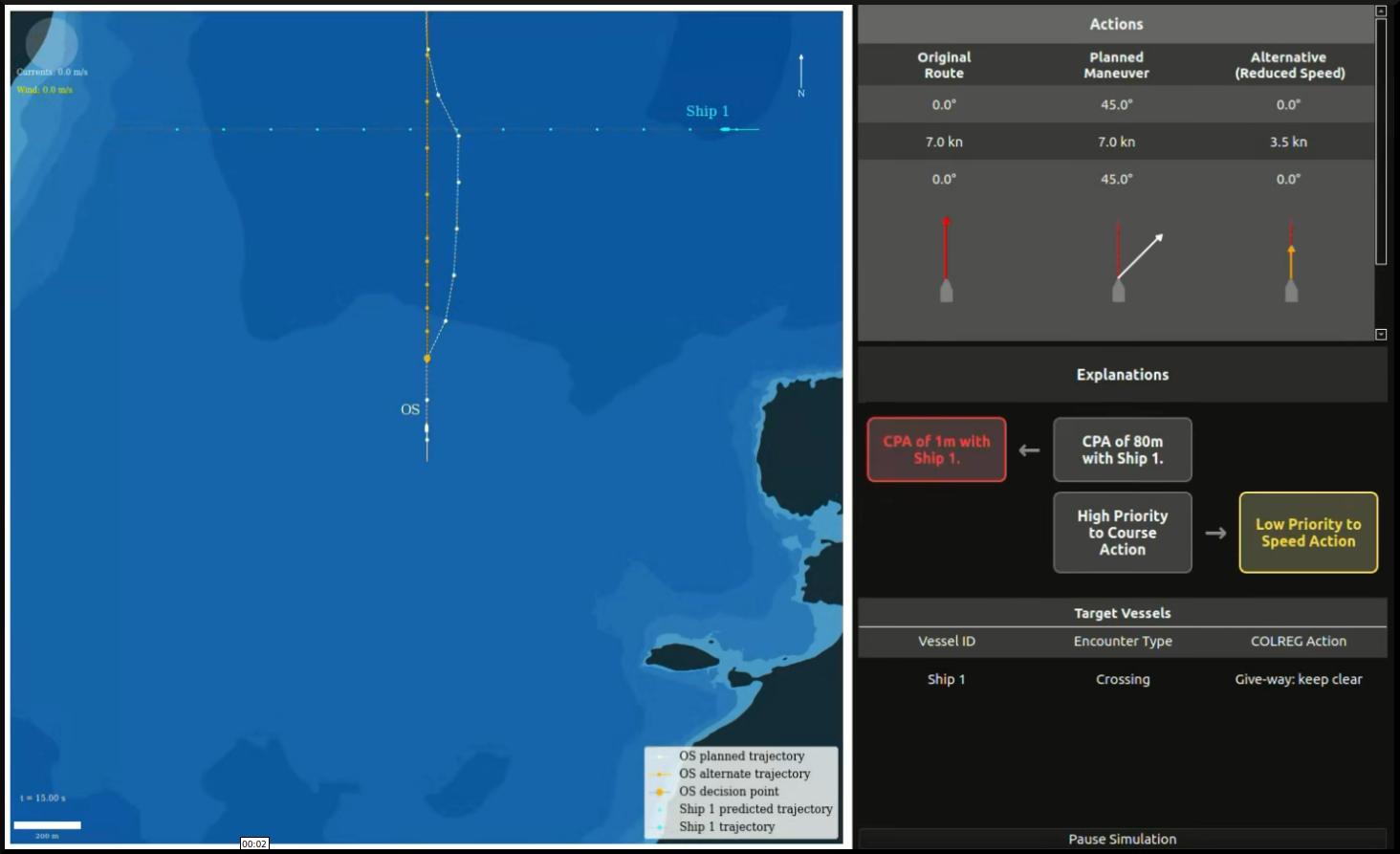} \\[1ex]
\includegraphics[width=0.45\textwidth]{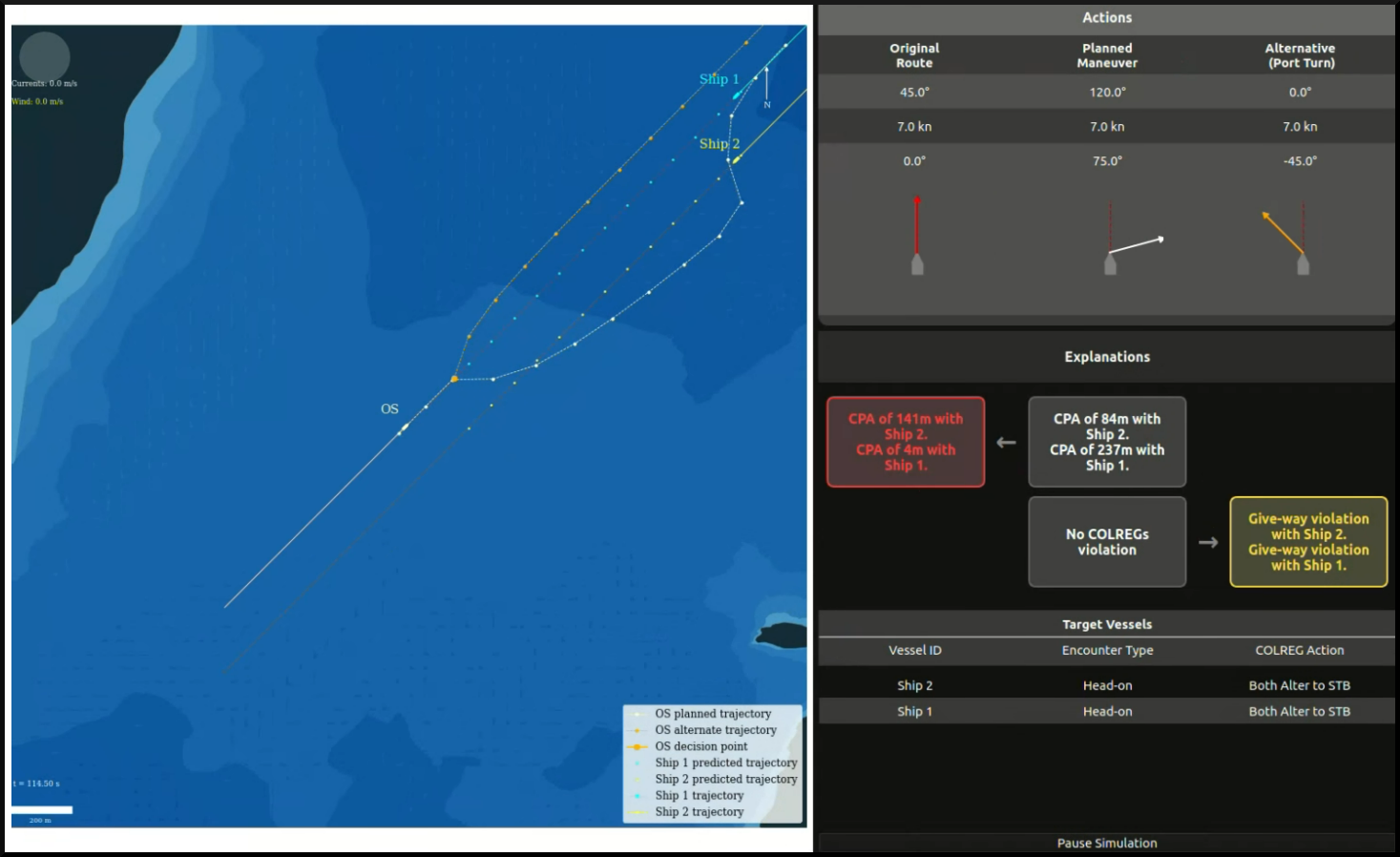} &
\includegraphics[width=0.45\textwidth]{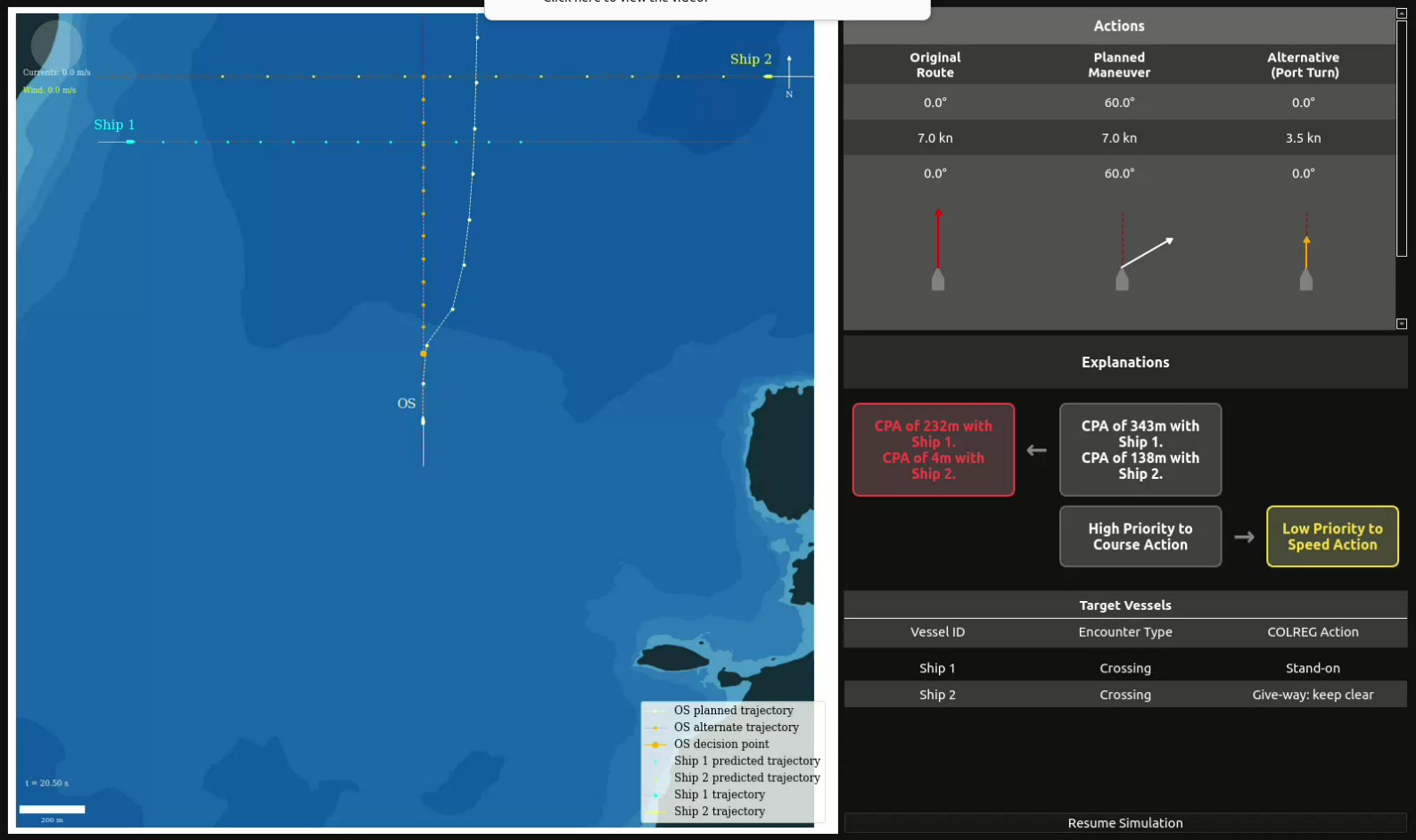}
\end{tabular}
\caption{Snapshots of traffic scenarios utilized in the user study, with the alternative maneuvers and associated explanations. Top right: Single-vessel, Head-on Encounter; Top left: Single-vessel, Crossing Encounter; Bottom left: Multi-vessel, two head-on encounters; Bottom Right: Multi-vessel, one crossing Give-Way, one crossing Stand-on.}
\label{fig:traffic_scenarios}
\end{figure*}

\section{User Study}\label{section:user_study}
We conducted an exploratory user study aimed at collecting feedback on the utility and preference towards contrastive explanations when monitoring a CAS. Since this paper implements a proof-of-concept of contrastive explanations, we aim for a sample size appropriate to obtain early-stage feedback, and consistent with related works (see e.g. \cite{veitch_operatorferryuserstudy2022, Lied2025}). Accordingly, we invited four experienced mariners holding (or having previously held) STCW-compliant deck officer certificates at the management level to participate in the user study.

Here, we outline the simulation framework before describing the methodology used in the user study. Afterwards, we cover the scenarios used for the user study and the questionnaire used to gather feedback.

\subsection{Simulation Framework}\label{section:simulator}

In this study, we use the collision avoidance simulation framework by Tengesdal et al. \cite{tengesdal_simulation_2023} for simulating the traffic scenarios. For the sake of simplicity, we focus only on traffic scenarios involving vessel-on-vessel encounters without grounding hazards, and assume that the CAS has perfect knowledge about the target vessel positions. The own-ship and target vessel models utilized here are based on the Telemetron vessel \cite{hagen_sbmpc_transition_2018}, an 8-meter long, 3-meter wide, highly maneuverable Rigid Buoyancy Boat (RBB) developed by Maritime Robotics, with maximum speed of upto 34 knots (17 m/s). Similar to related works with the SB-MPC algorithm \cite{hagen_sbmpc_transition_2018}, a kinematic representation of the model was used to generate candidate trajectories for evaluation. The autopilot used for obtaining references for the SB-MPC is a line-of-sight (LOS) guidance system \cite{breivikLOS_2008}. 

\subsection{Procedure}

After being briefed on the purpose and scope of the experiment, the participants are introduced to the notion of explainability and an overview of contrastive explanations. 
The participants are asked to imagine themselves in the role of a supervisor monitoring a CAS. They are informed about the own-ship and target vessel particulars and the region of operation for the trials. Then they are briefed about the experimental setup with each trial and asked to study the information provided and make a decision to accept or decline the proposed maneuver from the CAS.

Afterwards, a picture of the visualization interface is shown to familiarize the participants with all the information present. They are introduced to the relevant information indicated on the map, the CAS's actions with the associated contrastive explanations, and the perceived target vessels.

From thereon, the experiment commences, with the participant being presented with each trial, and their responses recorded. Once all the trials are complete, they are asked to fill out a questionnaire, after which the user study is finished.

\subsection{Traffic Scenarios and Experimental Trials}

Each user study lasts about 30-45 minutes, containing 4 trials each. For each trial, we prepared single or multi-vessel encounters occurring in calm conditions on a ferry route. We chose two single-vessel encounters and two multi-vessel encounters to gauge the participants' response to scenarios with varying complexity, as shown in \autoref{fig:traffic_scenarios}. For each scenario, we program the target vessels to move in a straight line and not perform avoidance maneuvers of their own.

In each trial, the participant is shown a video of a traffic scenario on the visualization interface. The video starts at some time ahead of the decision point where the CAS executes an avoidance maneuver. The task does not evaluate the participants’ performance or their ability to react quickly during supervision. Therefore, they are allowed to pause, play, and rewind the video as they wish. However, they are reminded of the time-critical nature of the situation. Once the participant is ready to make their decision, they are asked to record their answers on a sheet of paper. The procedure is then repeated for the subsequent trials.

\subsection{Questionnaire and User feedback}\label{section:questionnaire}

In order to study the navigators’ opinion regarding the utility, efficiency, and satisfaction in having contrastive explanations for collision avoidance, we provided the participants with a questionnaire consisting of six rating-based questions on a linear scale (from 0 to 10), where 0 indicates a bad rating, 10 indicates a good rating, and 5 indicates an uncertain or neutral response: 

\renewcommand{\theenumi}{Q\arabic{enumi})}
\renewcommand{\labelenumi}{Q\arabic{enumi})}
\begin{enumerate}
    \item How useful are contrastive explanations in the supervision of a CAS?
    \item How useful are contrastive explanations in understanding the system's reasoning?
    \item How useful is it to have an alternative maneuver (in addition to the original route) to compare with the proposed maneuver from the CAS?
    \item How sufficient is the information provided in the explanations in making the participants' decision? 
    \item How do contrastive explanations affect the participants' ability to react quickly to a scenario?
    \item What is the participants' overall preference towards contrastive explanations?
\end{enumerate}

Additionally, questions of a more qualitative nature, with free-text answers, were given to perform a semi-structured interview. We also added subjective questions to gain feedback and suggestions regarding the participants' preferences on the choice of alternatives, and when contrastive explanations should be provided. Some feedback was also obtained via conversation during the user study. 

\section{User Study Results}\label{section:results}

We present the key observations from the user study to reveal the participants' feedback and response towards contrastive explanations and to support our claims.

\subsection{Qualitative Results from the User Study}

\begin{figure}[t]
    \centering
    \includegraphics[width=0.95\linewidth]{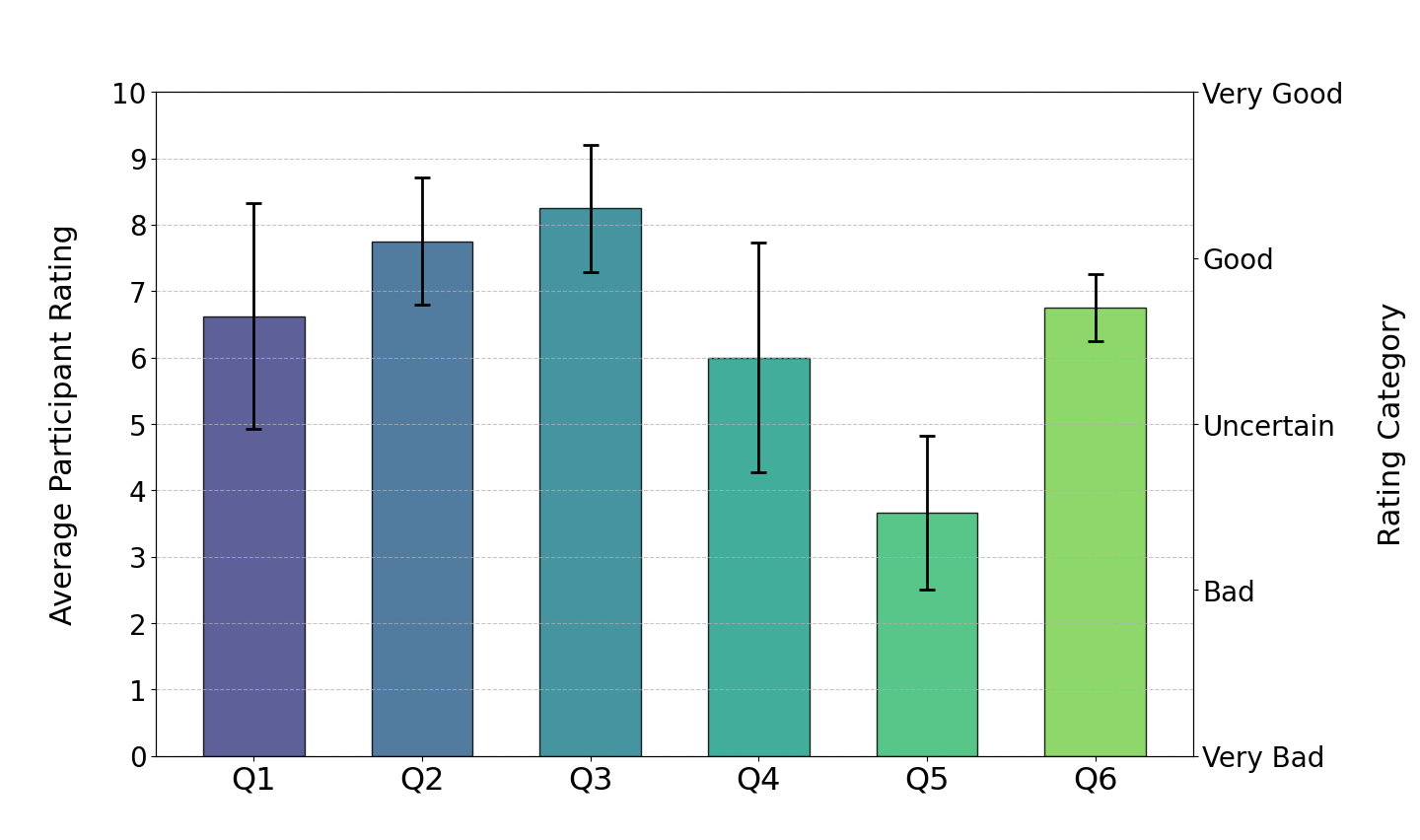}
    \caption{Responses from participants with management level deck-officer certificate to the questions in \autoref{section:questionnaire}. Participants had positive reviews of contrastive explanations when supervising a collision avoidance system, especially when comparing the proposed maneuver with alternative maneuvers.}
    \label{fig:user_ratings}
\end{figure}

We tabulate the responses of participants for the trials in \autoref{tab:participant_responses}, and the rating scores from the questionnaire in \autoref{section:questionnaire} in \autoref{fig:user_ratings}.

Responses to Q1 and Q6 in the questionnaire indicate that participants generally perceived the contrastive explanations as an effective and satisfactory tool for supervising a CAS. Answers to Q2 further support this view, suggesting that comparing maneuvers with accompanying explanations helped participants understand the system’s reasoning, thereby supporting our first claim. Furthermore, feedback on Q3 shows that participants found the availability of an alternative maneuver for comparison beneficial, thereby supporting our second claim. One participant emphasized the value of the contrastive alternatives, particularly in Trial 3, where it facilitated the decision to accept the proposed maneuver. The participants also provided comments regarding the choice of characteristics for the alternative, which we discuss in the next section.


However, responses to Q4 reveal some uncertainty regarding the sufficiency of the information provided in performing their task of accepting or rejecting a proposed maneuver, suggesting the need for further work to answer the third claim. One reason for this could be the design of the visualization interface itself, which lacks details that a navigator can normally expect, such as those from plotting tools like electronic chart display information system (ECDIS) or automatic radar plotting aid (ARPA), or live camera feed. However, the responses may also point to the potential value of a more advanced explanation framework, such as the use of conversational AI \cite{hodne_convAI}, which allows the supervisor to inquire about possible maneuvers in-depth.

Responses for Q5 reveal more workload during supervision, likely due to processing more information when explanations are provided \cite{van_de_merwe_influence_2024}. This effect can vary with scenario complexity. For example, one participant mentioned for the first scenario, \textit{'I didn't pay much attention to the explanations since I could look at the trajectory to understand'}. This indicates that for simpler scenarios, transparency information without explanations (such as in \cite{van_de_merwe_influence_2024, madsen_improving_2025}) is often enough to understand the vessel's reasoning behind avoidance maneuvers.
Conversely, another participant said, \textit{'contrastive explanations can be useful in complex scenarios involving multiple ships and can help reduce reaction times'}. This reflects the selective nature of contrastive explanations, revealing only those features that differentiate the proposed maneuver from the original route or an alternative. The comments from the participants, therefore, indicate that explanations need not be shown for all scenarios, and can be provided on demand, when they can be most utilized.

We acknowledge that our results may be influenced by the novelty effect, where positive responses arise from the newness of an innovation rather than its actual effectiveness \cite{elston_novelty_2021}. As this effect typically diminishes over time, the favorable perceptions of the explanations by the participants should be interpreted with caution, as they may partly reflect initial enthusiasm rather than genuine utility.

\begin{table}[h]
\centering
\begin{tabular}{c c c c c}
\hline
\textbf{Trial} & \textbf{P1} & \textbf{P2} & \textbf{P3} & \textbf{P4} \\
\hline
Trial 1 & Accept & Accept & Accept & Accept \\
Trial 2 & Accept & Accept & Accept & Accept \\
Trial 3 & Decline & Decline & Decline & Accept \\
Trial 4 & Accept & Decline & Accept & Accept \\
\hline
\end{tabular}
\caption{Participant responses across trials. A majority of participants declined the proposed maneuver in Trial 3 due to a need to communicate with other vessels. }
\label{tab:participant_responses}
\end{table}

\subsection{Additional Comments and Limitations}
For complex scenarios—such as Trial 3, three of four participants preferred to intervene because they wanted to communicate with the target vessels and choose an alternative maneuver that violates a COLREG rule (as shown in \autoref{fig:traffic_scenarios}, bottom left). This suggests the need for further studies on the effectiveness of explanations and human-AI interaction in CASs that communicate and collaborate with neighboring vessels (see e.g. \cite{AKDAG2022}).

One participant noted that the type of vessel affects the efficacy of contrastive explanations. Larger, high-inertia vessels have slower dynamics, making action choices more consequential. In such cases, comparing different trajectories helps reveal the system’s understanding, suggesting that contrastive explanations may be particularly valuable for high-inertia vessels.

The participants had mixed reviews as to when explanations should be displayed. This, combined with the varied effects of explanations across simple and complex scenarios, can indicate the benefits of a demand-driven approach \cite{saghafian_understanding_2025}, in which explanations are provided to the supervisor only when requested. We leave this as future work.

Regarding the choice of alternatives, participants suggested adding alternatives with different CPA distances and Time-to-CPA (TCPA). One participant noted, \textit{'Alternatives with speed reduction are not necessary, since it is rare to reduce speed during avoidance maneuvers, unless absolutely necessary'}. Three of four participants preferred to select the alternative themselves, with one suggesting that an intelligent system could provide a first option, allowing the supervisor to choose a different one if desired. Providing a list of scenario-based alternatives may improve interactivity and user satisfaction, but the effect on the supervisor’s ability to react should be monitored.  As one participant said, \textit{'the choice of alternative is very dependent on the scenario'}, consistent with claims in the literature \cite{miller_explanation_2019}. Thus, further studies towards scenario-based alternatives are warranted, potentially using data-driven methods.

\section{Conclusion \& Future Work}\label{section:conclusions}

In this study, we explored the use of contrastive explanations to support CAS supervision and proposed a framework for generating them in optimization-based collision avoidance planners. We implemented the framework using the SB-MPC planner and designed a visualization interface to convey the planner’s reasoning. An exploratory study with experienced mariners demonstrated that contrastive explanations are an effective and satisfactory method for supporting CAS supervision, aiding operators in understanding the system's rationale—particularly when highly relevant alternatives are used for comparison. Crucially, our findings revealed a practical trade-off: while contrastive explanations are highly valuable in complex, multi-vessel scenarios, they can increase cognitive workload in simpler encounters where baseline trajectory transparency is often sufficient.


Future work should expand upon these findings in several key directions to optimize human-AI teaming. First, to mitigate information overload, subsequent research should explore demand-driven explanation strategies, where operators can request explanations or a curated list of contextually relevant alternatives (e.g., varying CPA or TCPA) only when needed. Second, addressing the participants' need for more comprehensive situational data, future interfaces should integrate standard navigational overlays (such as ECDIS and ARPA) or explore conversational AI to allow supervisors to query the CAS's maneuver planning in greater depth. Third, the framework could be integrated into advanced CASs that actively communicate and negotiate intents with target vessels. Finally, to rigorously validate these findings, contrastive explanations must be evaluated against other explanation modalities through comprehensive, scenario-based user trials in high-fidelity nautical simulators or Remote Operations Centers (ROCs).

\bibliographystyle{unsrt}
\bibliography{references}

\begin{thebibliography}{10}

\bibitem{kurt_asvimpacts_2022}
Ismail Kurt and Murat Aymelek.
\newblock Operational and economic advantages of autonomous ships and their perceived impacts on port operations.
\newblock {\em Maritime Economics \& Logistics}, 24, 2022.

\bibitem{vagale_path_2021}
Anete Vagale, Robin Bye, Rachid Oucheikh, Ottar Osen, and Thor Fossen.
\newblock Path planning and collision avoidance for autonomous surface vehicles ii: a comparative study of algorithms.
\newblock {\em Journal of Marine Science and Technology}, 2021.

\bibitem{rodseth_towards_2022}
Ørnulf~Jan Rødseth, Lars~Andreas Lien~Wennersberg, and Håvard Nordahl.
\newblock Towards approval of autonomous ship systems by their operational envelope.
\newblock {\em Journal of Marine Science and Technology}, 27(1):67--76, 2022.

\bibitem{madsen_improving_2025}
Andreas~Nygard Madsen, Andreas Brandsæter, Koen van~de Merwe, and Jooyoung Park.
\newblock Improving decision transparency in autonomous maritime collision avoidance.
\newblock {\em Journal of Marine Science and Technology}, 2025.

\bibitem{endsley_supporting_2023}
Mica~R. Endsley.
\newblock Supporting {Human}-{AI} {Teams}:{Transparency}, explainability, and situation awareness.
\newblock {\em Computers in Human Behavior}, 140:107574, 2023.

\bibitem{van_de_merwe_influence_2024}
Koen Van De~Merwe, Steven Mallam, Salman Nazir, and Øystein Engelhardtsen.
\newblock The {Influence} of {Agent} {Transparency} and {Complexity} on {Situation} {Awareness}, {Mental} {Workload}, and {Task} {Performance}.
\newblock {\em Journal of Cognitive Engineering and Decision Making}, 18(2):156--184, 2024.

\bibitem{endsley_toward_1995}
Mica Endsley.
\newblock Toward a {Theory} of {Situation} {Awareness} in {Dynamic} {Systems}.
\newblock {\em Human Factors: The Journal of the Human Factors and Ergonomics Society}, 37:32--64, 1995.

\bibitem{psw_model}
Raja Parasuraman, Thomas Sheridan, and Christopher Wickens.
\newblock A model for types and levels of human interaction with automation. ieee trans. syst. man cybern. part a syst. hum. 30(3), 286-297.
\newblock {\em IEEE transactions on systems, man, and cybernetics. Part A, Systems and humans : a publication of the IEEE Systems, Man, and Cybernetics Society}, 30:286--97, 06 2000.

\bibitem{brandsaeter_simulator-based_2024}
Andreas Brandsæter and Andreas Madsen.
\newblock A simulator-based approach for testing and assessing human supervised autonomous ship navigation.
\newblock {\em Journal of Marine Science and Technology}, 29(2):432--445, 2024.

\bibitem{Omeiza2022ExplanationsSurvey}
Daniel Omeiza, Helena Webb, Marina Jirotka, and Lars Kunze.
\newblock {Explanations in Autonomous Driving: A Survey}.
\newblock {\em IEEE Transactions on Intelligent Transportation Systems}, 23(8):10142--10162, 2022.

\bibitem{gjaerum_real-time_2023}
Vilde~B. Gjærum, Inga Strümke, Anastasios~M. Lekkas, and Timothy Miller.
\newblock Real-{Time} {Counterfactual} {Explanations} {For} {Robotic} {Systems} {With} {Multiple} {Continuous} {Outputs}.
\newblock {\em IFAC-PapersOnLine}, 56(2):7--12, 2023.
\newblock Publisher: Elsevier BV.

\bibitem{gjaerum_explaining_2021}
Vilde~B Gjaerum, Inga Strümke, Ole~Andreas Alsos, and Anastasios~M Lekkas.
\newblock Explaining a {Deep} {Reinforcement} {Learning} {Docking} {Agent} {Using} {Linear} {Model} {Trees} with {User} {Adapted} {Visualization}.
\newblock {\em Marine Science and Engineering}, 2021.

\bibitem{singh_fuzzyxai2023}
Vijander Singh, Ottar~L. Osen, and Robin~T. Bye.
\newblock Explainable artificial intelligence for autonomous surface vessels by fuzzy-based collision avoidance system.
\newblock In Tomonobu Senjyu, Chakchai So-In, and Amit Joshi, editors, {\em Smart Trends in Computing and Communications}, Singapore, 2023. Springer Nature Singapore.

\bibitem{veitch_human-centered_2021}
Erik Veitch and Ole~Andreas Alsos.
\newblock Human-centered explainable artificial intelligence for marine autonomous surface vehicles.
\newblock {\em Journal of Marine Science and Engineering}, 9(11), 2021.

\bibitem{Lipton1990}
Peter Lipton.
\newblock Contrastive explanation.
\newblock {\em Royal Institute of Philosophy Supplement}, 27:247--266, 1990.

\bibitem{miller_explanation_2019}
Tim Miller.
\newblock Explanation in artificial intelligence: {Insights} from the social sciences.
\newblock {\em Artificial Intelligence}, 267:1--38, 2019.

\bibitem{option_centric_Luo2019}
Ruikun Luo, Na~Du, Kevin~Y. Huang, and X.~Jessie Yang.
\newblock Enhancing transparency in human-autonomy teaming via the option-centric rationale display.
\newblock {\em Proceedings of the Human Factors and Ergonomics Society Annual Meeting}, 63(1):166--167, 2019.

\bibitem{gros_scaffolding_2023}
André Groß, Amit Singh, Ngoc~Chi Banh, Birte Richter, Ingrid Scharlau, Katharina~J. Rohlfing, and Britta Wrede.
\newblock Scaffolding the human partner by contrastive guidance in an explanatory human-robot dialogue.
\newblock {\em Frontiers in Robotics and AI}, 10, 2023.
\newblock Publisher: Frontiers.

\bibitem{krarup_contrastive_2021}
Benjamin Krarup, Senka Krivic, Daniele Magazzeni, Derek Long, Michael Cashmore, and David~E. Smith.
\newblock Contrastive {Explanations} of {Plans} through {Model} {Restrictions}.
\newblock {\em Journal of Artificial Intelligence Research}, 72:533--612, 2021.

\bibitem{omeiza_towards_2021}
Daniel Omeiza, Helena Web, Marina Jirotka, and Lars Kunze.
\newblock Towards {Accountability}: {Providing} {Intelligible} {Explanations} in {Autonomous} {Driving}.
\newblock In {\em 2021 {IEEE} {Intelligent} {Vehicles} {Symposium} ({IV})}, 2021.

\bibitem{brandao_towards_2021}
Martim Brandão, Gerard Canal, Senka Krivić, and Daniele Magazzeni.
\newblock Towards providing explanations for robot motion planning.
\newblock In {\em 2021 {IEEE} {International} {Conference} on {Robotics} and {Automation} ({ICRA})}, 2021.
\newblock ISSN: 2577-087X.

\bibitem{schneider_ce-mrs_2024}
Ethan Schneider, Daniel Wu, Devleena Das, and Sonia Chernova.
\newblock {CE}-{MRS}: {Contrastive} {Explanations} for {Multi}-{Robot} {Systems}.
\newblock {\em IEEE Robotics and Automation Letters}, 9(11):10121--10128, 2024.
\newblock arXiv:2410.08408 [cs].

\bibitem{AKDAG2022}
Melih Akdağ, Petter Solnør, and Tor~Arne Johansen.
\newblock Collaborative collision avoidance for maritime autonomous surface ships: A review.
\newblock {\em Ocean Engineering}, 2022.

\bibitem{van_de_merwe_towards2023}
Koen van~de Merwe, Steven Mallam, Øystein Engelhardtsen, and Salman Nazir.
\newblock Towards an approach to define transparency requirements for maritime collision avoidance.
\newblock {\em Proceedings of the Human Factors and Ergonomics Society Annual Meeting}, 67(1):483--488, September 2023.

\bibitem{johansen_ship_2016}
Tor~Arne Johansen, Tristan Perez, and Andrea Cristofaro.
\newblock Ship {Collision} {Avoidance} and {COLREGS} {Compliance} {Using} {Simulation}-{Based} {Control} {Behavior} {Selection} {With} {Predictive} {Hazard} {Assessment}.
\newblock {\em IEEE Transactions on Intelligent Transportation Systems}, 17(12):3407--3422, 2016.

\bibitem{hagen_scenario-based_2022}
I.B. Hagen, D.K.M. Kufoalor, T.A. Johansen, and E.F. Brekke.
\newblock Scenario-{Based} {Model} {Predictive} {Control} with {Several} {Steps} for {COLREGS} {Compliant} {Ship} {Collision} {Avoidance}.
\newblock {\em IFAC-PapersOnLine}, 55(31):307--312, 2022.

\bibitem{heuillet_explainability_2021}
Alexandre Heuillet, Fabien Couthouis, and Natalia Díaz-Rodríguez.
\newblock Explainability in deep reinforcement learning.
\newblock {\em Knowledge-Based Systems}, 214, 2021.

\bibitem{hodne_convAI}
Philip Hodne, Oskar~K. Skåden, Ole~Andreas Alsos, Andreas Madsen, and Thomas Porathe.
\newblock Conversational user interfaces for maritime autonomous surface ships.
\newblock {\em Ocean Engineering}, 310:118641, 2024.

\bibitem{veitch_operatorferryuserstudy2022}
Erik Veitch, Kim Christensen, Markus Log, Erik Valestrand, Sigurd Hilmo~Lundheim, Martin Nesse, Ole Alsos, and Martin Steinert.
\newblock From captain to button-presser: operators’ perspectives on navigating highly automated ferries.
\newblock {\em Journal of Physics: Conference Series}, 2311:012028, 07 2022.

\bibitem{Lied2025}
Henrik~Viken Lied, Taufik~Akbar Sitompul, and Ole~Andreas Alsos.
\newblock Redesigning ui for teleoperating rov using apple vision pro: A study on usability and workload.
\newblock In {\em Proceedings of the 20th International Conference on Virtual Reality Continuum and Its Applications in Industry}. ACM, December 2025.

\bibitem{tengesdal_simulation_2023}
Trym Tengesdal and Tor~A. Johansen.
\newblock Simulation {Framework} and {Software} {Environment} for {Evaluating} {Automatic} {Ship} {Collision} {Avoidance} {Algorithms}.
\newblock In {\em 2023 {IEEE} {Conference} on {Control} {Technology} and {Applications} ({CCTA})}, Bridgetown, Barbados, 2023. IEEE.

\bibitem{hagen_sbmpc_transition_2018}
I.~B. Hagen, D.~K.~M. Kufoalor, E.~F. Brekke, and T.~A. Johansen.
\newblock Mpc-based collision avoidance strategy for existing marine vessel guidance systems.
\newblock In {\em 2018 IEEE International Conference on Robotics and Automation (ICRA)}, 2018.

\bibitem{breivikLOS_2008}
Morten Breivik and Thor~I. Fossen.
\newblock Guidance laws for planar motion control.
\newblock In {\em 2008 47th IEEE Conference on Decision and Control}, pages 570--577, 2008.

\bibitem{elston_novelty_2021}
Dirk~M. Elston.
\newblock The novelty effect.
\newblock {\em Journal of the American Academy of Dermatology}, 85:565--566, 2021.

\bibitem{saghafian_understanding_2025}
Mina Saghafian, Dorthea Mathilde~Kristin Vatn, Stine~Thordarson Moltubakk, Lene~Elisabeth Bertheussen, Felix~Marcel Petermann, Stig~Ole Johnsen, and Ole~Andreas Alsos.
\newblock Understanding automation transparency and its adaptive design implications in safety–critical systems.
\newblock {\em Safety Science}, 184:106730, 2025.

\end{thebibliography}

\end{document}